# Stochastic Parrots or ICU Experts? Large Language Models in Critical Care Medicine: A Scoping Review

**Tongyue Shi[1,2,3,4], Jun Ma[1,5], Zihan Yu[6], Haowei Xu[1,4], Minqi Xiong[7], Meirong Xiao[1,2], Yilin Li[8], Huiying Zhao[9], and Guilan Kong[1,2,3,4]\***

[1] National Institute of Health Data Science, Peking University, Beijing, China

[2] Institute of Medical Technology, Peking University Health Science Center, Beijing, China

[3] Institute for Artificial Intelligence, Peking University, Beijing, China

[4] Advanced Institute of Information Technology, Peking University, Hangzhou, China

[5] Peking University Third Hospital, Beijing, China

[6] Department of Computer Science, University of Liverpool, Liverpool, UK

[7] Johns Hopkins School of Medicine, Baltimore, MD, USA

[8] Fielding School of Public Health, University of California, Los Angeles, CA, USA

[9] Department of Critical Care Medicine, Peking University People's Hospital, Beijing, China

\* Address correspondence to: guilan.kong@hsc.pku.edu.cn

## Abstract

With the rapid development of artificial intelligence (AI), large language models (LLMs) have shown strong capabilities in natural language understanding, reasoning, and generation, attracting amounts of research interest in applying LLMs to health and medicine. Critical care medicine (CCM) provides diagnosis and treatment for critically ill patients who often require intensive monitoring and interventions in intensive care units (ICUs). Can LLMs be applied to CCM? Are LLMs just like stochastic parrots or ICU experts in assisting clinical decision-making? This scoping review aims to provide a panoramic portrait of the application of LLMs in CCM. Literature in seven databases, including PubMed, Embase, Scopus, Web of Science, CINAHL, IEEE Xplore, and ACM Digital Library, were searched from January 1, 2019, to June 10, 2024. Peer-reviewed journal and conference articles that discussed the application of LLMs in critical care settings were included. Studies were excluded if they did not address LLMs in CCM or were non-English publications. From an initial 619 articles, 24 were selected for final review using a standard scoping review methodology. After a rigorous examination, this review grouped applications of LLMs in CCM into three categories: clinical decision support, medical documentation and reporting, and medical education and doctor-patient communication. Compared to traditional AI models, LLMs have advantages in handling unstructured data and do not require manual feature engineering. Meanwhile, applying LLMs to CCM faces challenges, including hallucinations and poor interpretability, sensitivity to prompts, bias and alignment challenges, and privacy and ethics issues. Future research should enhance model reliability and interpretability, improve training and deployment scalability, integrate up-to-date medical knowledge, and strengthen privacy and ethical guidelines. As LLMs evolve, they could become key tools in CCM to help improve patient outcomes and optimize healthcare delivery. This study is the first review of LLMs in CCM, aiding researchers, clinicians, and policymakers to understand the current status and future potentials of LLMs in CCM.

# 1. Introduction

Critical care medicine (CCM), also called intensive care medicine, is an essential field dedicated to the management of severely ill patients, emphasizing rapid and life-critical decision-making and interventions. CCM deals with patients who have severe conditions and injuries such as sepsis, acute kidney injury (AKI), and acute respiratory distress syndrome (ARDS), potentially leading to a deteriorative state in the intensive care units (ICUs) [1]. Sepsis accounted for approximately 11 million deaths in 2017, making up about 20% of all global deaths [2]. The incidence of AKI in the ICU could reach up to 66% globally [3]. Among those who received renal replacement therapy, some of the most critically ill individuals in the ICU, the mortality rate was approximately 50% [3]. Recent studies [4, 5] found that the incidence of ARDS in the ICU was about 10%, and the ICU mortality of ARDS was approximately 35% in high-income countries. While in resource-limited settings, the ICU mortality of ARDS could be as high as 50% due to the disparities in healthcare quality [4]. The aging of the population and the deterioration of the living environment continue to pose new challenges to human health [6, 7], and there is a substantial rise in the demand for intensive care services [8]. The physicians and nurses in ICUs need to deal with large amounts of patient data and maintain high efficiency under high pressure [4, 9]. Critical care's dynamic and severe nature demands intelligent decision-support tools that can help physicians improve diagnostic accuracy, optimize therapeutic strategies, and provide timely clinical decision-making.

Artificial intelligence (AI) technologies, especially generative AI models, have developed rapidly in recent years. The advent of large language models (LLMs), such as those based on the Transformer architecture [10] and pre-trained on extensive text corpora, has marked a substantial advancement in natural language processing (NLP). With billions of parameters, these LLMs have demonstrated remarkable capabilities in understanding and generating human-like text [11]. LLMs have been implemented in different contexts, such as answering questions, summarizing texts, and participating in open-domain conversations [12]. Among these LLMs, OpenAI's ChatGPT [11] has become a focal point since its launch in November 2022. Originating from the GPT series, this AI-driven chatbot utilizes a blend of supervised and reinforcement learning strategies. Its rapid adoption indicates the growing curiosity and reliance on such technologies to streamline communication and decision-making processes. With the advancement of AI technologies, OpenAI then quickly introduced upgraded versions of ChatGPT, GPT-4, GPT-4o, and GPT-4o mini in 2023 and 2024, offering enhanced multimodal capabilities to handle diverse inputs like text, images, and table files.

Despite the origins of LLM models not being directly tied to health and medicine, the flexibility of LLMs has allowed them to become valuable assets in medical settings, providing support for diagnostic assistance, medical professional training, and medical summarization [13]. In the field of CCM, the emergence of LLM demonstrates its unique potential. Similar to the application of LLMs in informing cancer patients of diagnosis, treatment methods, and side effects [14], LLMs in CCM can help make life-or-death decisions after fusing large volumes of patient data in a short time [15]. Physicians in CCM face enormous workloads and pressure, involving LLMs in different clinical decision-making scenarios in CCM will help reduce the workload of physicians and improve healthcare quality. Recent studies [16-19] have explored the potential applications of LLMs in health and medicine, revealing their capabilities to assist junior physicians in medical diagnostics and decision-making. Compared to human practitioners, LLMs have been perceived as more understanding and efficient[20]. However, LLMs face challenges when applied in medicine, such as uncertain accuracy and coherence, recency bias, hallucinations, poor interpretability, and ethical issues [17]. Among them, hallucinations are one of the biggest drawbacks of LLMs, which make them act like stochastic parrots [21].

This study aims to systematically review the applications of LLMs in CCM, identifying the advantages, challenges, and future potentials of LLMs in this field. By exploring how these advanced models have been applied in CCM to date, this review seeks to identify the gaps in current research and examine whether LLMs can truly enhance clinical decision-making and improve patient outcomes in the ICUs. This review is important in moving beyond the hype and assessing the real-world studies of employing LLMs in CCM.

Three key research questions were designed to be answered by this review. (1) What is the current status of LLM applications within the critical care setting? (2) What are the recognized advantages and challenges of utilizing LLMs in CCM? (3) What research directions should be taken in the future to promote the application of LLMs in CCM? By addressing the above three questions, this review endeavors to provide a clear portrait of and identify the research gap in the applications of LLMs in CCM, analyzing whether they are just stochastic parrots that can mimic human responses based on probability calculation or emerging ICU experts capable of providing timely and highly personalized diagnosis and treatment recommendations. Through this comprehensive review, we aim to outline a roadmap for future research and implementation of LLMs in CCM that could enable them to transform critical care effectively.

## 2. Overview of LLMs in Health and Medicine

LLMs have transformed numerous fields through their unprecedented capabilities in understanding and generating natural language. Generally, LLMs refer to Transformer-based language models containing hundreds of billions or more parameters, trained on vast amounts of text data. Typical examples include GPT-3 [22], PaLM [23], and LLaMA [24].

### 2.1 Evolution of LLMs

The evolution of LLMs represents a complex and progressive journey intertwined with the advancements in generative models, sequence models, and pre-trained language models. The developing course of LLMs is shown in Figure 1.

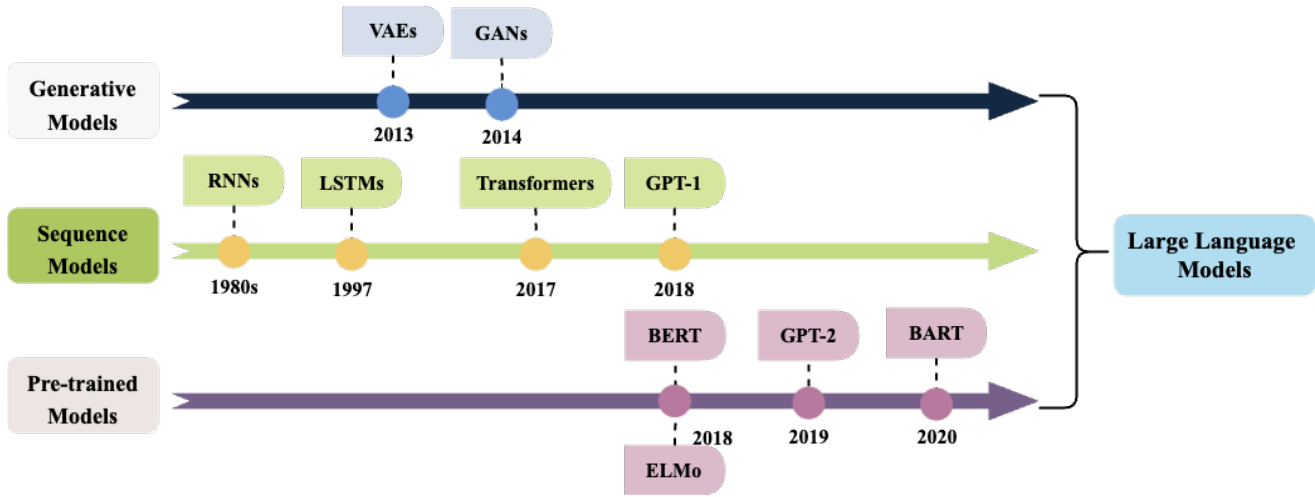

**Figure 1. The developing course of large language models.**

**Generative Models** Initially, research in this domain concentrated on generative models [25] for generating text [26], images [27], audio [28], and other AI-generated content [29] that closely resemble the training data. Generative adversarial networks (GANs) [30] and variational autoencoders (VAEs) [31] were typical generative models for text generation, machine translation, and image synthesis. These models aim to generate new examples with a similar distribution as the training set. However, their performance may be constrained by the availability of large-scale training datasets and the computational resources required for model training.

**Sequence Models** Sequence models operate on the principle that each word or token in a sequence can be predicted based on the preceding tokens [32]. The term "sequence" refers to the method of processing and predicting data sequences. The Transformer architecture is an important milestone in the development of sequence models. It relies on a self-attention mechanism, allowing the model to weigh the importance of different words in each sequence to make predictions. Compared to recurrent neural networks (RNNs) [33] or long short-term memory networks (LSTMs) [34], Transformers can capture dependencies across the entire sequence more effectively, and they have become the backbone of many state-of-the-art LLMs.

**Pre-trained Language Models** As an early attempt, ELMo [35] captured context-aware word representations by pre-training a bidirectional LSTM network and subsequently fine-tuning it for specific downstream tasks. BERT [36] was developed based on the highly parallelizable Transformer architecture with self-attention mechanisms and pre-trained by a bidirectional language model on a large-scale unlabeled corpus. A "pre-training and fine-tuning" learning paradigm was developed in pre-trained language models (PLMs) and has inspired extensive follow-up research, which introduces different architectures such as GPT-2 [37] and BART [38]. This paradigm typically requires fine-tuning PLMs to adapt them to various downstream tasks.

**Large Language Models** Researchers have found that scaling PLMs can generally improve performance on downstream tasks following the scaling laws [39]. Some studies [22, 23] have explored the performance limits by training increasingly larger PLMs, such as GPT-3 with 175 billion parameters and PaLM with 540 billion parameters. These large-scale PLMs exhibit different behaviors from smaller PLMs like BERT with 330 million parameters and GPT-2 with 1.5 billion parameters. For instance, GPT-3 can address few-shot learning tasks through in-context learning, whereas GPT-2 performs poorly for similar tasks [22]. Consequently, the research community termed these large-scale PLMs as LLMs. As a typical application of LLMs, ChatGPT leverages the GPT series of LLMs for conversational purposes, demonstrating impressive human-like dialogue capabilities. Solving complex tasks which involve multiple steps is challenging for LLMs [40]. Nevertheless, prompting strategies such as chain-of-thought (CoT) [41], which include intermediate reasoning steps, can help tackle complex arithmetic, commonsense, and symbolic reasoning tasks.

## 2.2 Applications of LLMs in Health and Medicine

LLMs exhibit substantial potential in different medical decision-making scenarios, including clinical decision support, medical document summarization, doctor-patient communication, and medical research. In clinical practice, LLMs can be utilized to provide supplemental treatments and diagnoses across different departments, such as internal medicine [13], surgery [42, 43], radiology [44, 45], and ophthalmology [46, 47]. The capabilities of summarizing and rephrasing information enable LLMs to generate detailed discharge summaries [48], radiology reports [49, 50], and other related medical documents, thereby reducing physicians' administrative burden. Furthermore, LLMs can automatize the international classification of diseases (ICD) coding process by extracting medical terms from clinical notes and assigning corresponding ICD codes, helping to improve coding efficiency and accuracy [51, 52]. The strong natural language understanding and generation capabilities of LLMs enable them to answer questions from patients with prostate cancer [53], nasal diseases [54], and liver cirrhosis [55], and can also provide emotional support to patients or caregivers [56]. ChatGPT has shown greater empathy than doctors when responding to patient inquiries [57]. In medical research, LLMs can serve as tools for literature retrieval and analysis [58], drug design and discovery [59], medical image segmentation [60], and medical language translation [61]. Utilizing LLMs for literature review and data analysis can help accelerate the research progress. Overall, applying LLMs in medicine can help improve clinical efficiency, support doctor-patient communication, and accelerate research progress to advance health and medicine.

# 3. Methods

## 3.1 Study Design

This study was conducted using a scoping review methodology, which is particularly well-suited for fields where the research topics are complex and varied. The methodology followed the structured five-stage approach proposed by Arksey and O'Malley [62, 63], which encompasses *Stage 1, Identifying the Research Question; Stage 2, Identifying Relevant Studies; Stage 3, Study Selection; Stage 4, Charting the Data;* and *Stage 5, Collating, Summarizing, and Reporting Results.* This study aimed to provide a comprehensive review of the application of LLMs in CCM.

## 3.2 Literature Search Strategy

To comprehensively investigate the applications of LLMs in CCM, we conducted the literature search across seven databases, including PubMed, Embase, Scopus, Web of Science, CINAHL, IEEE Xplore, and ACM Digital Library, for *Stage 2 Identifying Relevant Studies*. The time frame for our search spanned more than five years, from January 1, 2019, to June 10, 2024, encompassing the period since the emergence of LLMs in the field.

Our search strategy was meticulously designed by synthesizing keywords related to LLMs and CCM [18, 64-67]. Keywords related to large models included "Large Language Model", "LLM", "Generative Pre-trained Transformer", "GPT", "Generative Artificial Intelligence", and "Generative AI". For CCM, the keywords included "Critical Care", "Intensive Care Units", "Critical Illness", "Intensive Care", and "ICU". All these terms were combined using the "OR" and "AND" logical operators to ensure the retrieval of literature that addresses both research areas. The detailed search terms for each database are provided in Table 1.

**Table 1: Search terms used across multiple databases for the scoping review.**

| Database | Search String |
|---|---|
| PubMed | ("Critical Care"[MeSH] OR "Intensive Care Units"[MeSH] OR "Critical Illness"[MeSH] OR "Critical Care"[TIAB] OR "Intensive Care"[TIAB] OR "Critical Illness"[TIAB] OR "ICU"[TIAB]) AND ("Large Language Model"[TIAB] OR "LLM"[TIAB] OR "Generative Pre-trained Transformer"[TIAB] OR "GPT"[TIAB] OR "Generative Artificial Intelligence"[TIAB] OR "Generative AI"[TIAB]) |
| Scopus | TITLE-ABS-KEY ("Critical Care" OR "Intensive Care Units" OR "Critical Illness" OR "Intensive Care" OR "ICU") AND TITLE-ABS-KEY ("Large Language Model" OR "LLM" OR "Generative Pre-trained Transformer" OR "GPT" OR "Generative Artificial Intelligence" OR "Generative AI") |
| Web of Science | TS=("Critical Care" OR "Intensive Care Units" OR "Critical Illness" OR "Intensive Care" OR "ICU") AND TS=("Large Language Model" OR "LLM" OR "Generative Pre-trained Transformer" OR "GPT" OR "Generative Artificial Intelligence" OR "Generative AI") |
| Embase | ('Critical Care'/exp OR 'Intensive Care Units'/exp OR 'Critical Illness'/exp OR 'Intensive Care'/exp OR 'ICU'/exp OR 'Critical Care' OR 'Critical Illness' OR 'Intensive Care' OR 'ICU') AND ('Large Language Model' OR 'LLM' OR 'Generative Pre-trained Transformer' OR 'GPT' OR 'Generative Artificial Intelligence' OR 'Generative AI') |
| CINAHL | (MH "Critical Care" OR MH "Intensive Care Units" OR MH "Critical Illness" OR TI "Critical Care" OR TI "Intensive Care" OR TI "Critical Illness" OR TI "ICU" OR AB "Critical Care" OR AB "Intensive Care" OR AB "Critical Illness" OR AB "ICU") AND (TI "Large Language Model" OR AB "Large Language Model" OR TI "LLM" OR AB "LLM" OR TI "Generative Pre-trained Transformer" OR AB "Generative Pre-trained Transformer" OR TI "GPT" OR AB "GPT" OR TI "Generative Artificial Intelligence" OR AB "Generative Artificial Intelligence" OR TI "Generative AI" OR AB "Generative AI") |
| IEEE Xplore | ('Critical Care' OR 'Intensive Care Units' OR 'Critical Illness' OR 'Intensive Care' OR 'ICU') AND ('Large Language Model' OR 'LLM' OR 'Generative Pre-trained Transformer' OR 'GPT' OR 'Generative Artificial Intelligence' OR 'Generative AI') |

| ACM Digital Library | [[All: "critical care"] OR [All: "intensive care unit"] OR [All: "critical illness"] OR [All: "intensive care"] OR [All: "icu"]] AND [[All: "large language model"] OR [All: "llm"] OR [All: "generative pre-trained transformer"] OR [All: "gpt"] OR [All: "generative artificial intelligence"] OR [All: "generative ai"]] |
| --- | --- |

## 3.3 Study Selection

The process of *Stage 3 Select Studies* in this scoping review was conducted to ensure comprehensive coverage and relevance of the included literature. We undertook the study selection under the Preferred Reporting Items for Systematic Reviews and Meta-Analyses (PRISMA) framework [68].

**Inclusion Criteria**: In the first phase of the study selection, literature was included based on the following criteria: (1) relevant to LLMs and CCM. These studies explicitly used or commented on LLMs relevant to the field of CCM. (2) original research papers from peer-reviewed journals and conferences. (3) written in English.

**Exclusion Criteria**: Studies were excluded from the review if they met any of the following conditions: (1) irrelevant to LLMs or CCM. These studies did not focus on applying LLMs within the realm of CCM. (2) conference abstracts, preprint articles, books, patents, editorials, and review papers. (3) non-English literature.

The initial literature screening involved a review of titles, abstracts, and keywords by two independent reviewers (T.S. and Z.Y.). This first step was designed to eliminate irrelevant articles based on the inclusion and exclusion criteria. Articles that passed this preliminary filter were subjected to a more detailed full-text review. The same reviewers thoroughly checked the full articles during this second phase to confirm their eligibility. Discrepancies between reviewers at any stage of the selection process were resolved through discussion. A third reviewer (G.K.) was consulted to make the final decision if a consensus could not be reached.

## 3.4 Keyword Co-occurrence Network Analysis

Keyword co-occurrence network analysis [69] is a bibliometric method used to explore the relationships between keywords in academic papers. It involves constructing a network where nodes represent keywords and edges represent the co-occurrence of these keywords within the studied documents. It helps to identify the main research themes, trends, and potential research gaps by analyzing the frequency and patterns of keyword co-occurrences. This study used the VOSviewer (v1.6.20) software [70] to construct the bibliometric network using a clustering algorithm based on the visualization of similarity (VOS) method [71]. The software automatically extracts Keywords from a publication's title, abstract, or author-supplied keyword list. The frequency of co-occurrences of two keywords is determined by the number of publications in which both keywords appear together in the title, abstract, or keyword list. The VOS method starts by calculating the similarity between publications' keywords based on co-occurrence. Finally, a matrix used to arrange keywords spatially is then constructed according to their mutual similarities.

# 4. Results

## 4.1 Literature Search Results

This scoping review covered publications in the seven major databases from January 1, 2019, to June 10, 2024, and initially retrieved 619 articles. The study selection process is presented in Figure 2.

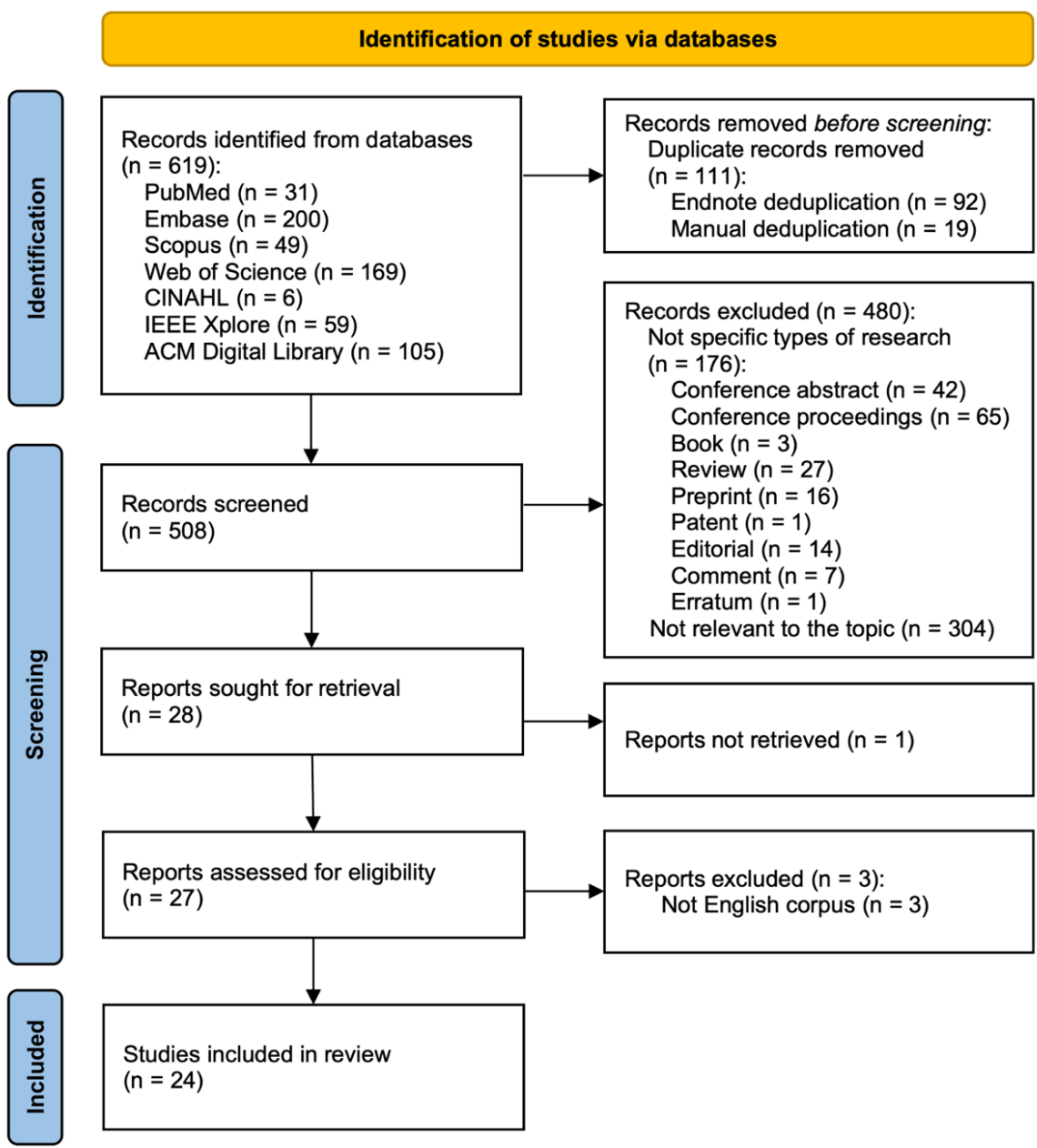

**Figure 2. PRISMA flowchart for study selection and quality assessment.**

The application of LLMs in CCM is a relatively innovative field, but research is still lacking, and the overall number of articles is relatively small. Finally, 24 articles met all the inclusion criteria and were chosen for this review. The research contents and publication details of the included studies are documented in Table 2.

**Table 2. Research Contents and publication details of the included studies in this review.**

| Authors | Published Year | Article Type | Journal or Conference Name | Country | Models | Research Contents |
|---|---|---|---|---|---|---|
| Bushuven et al. [72] | 2023 | Original Research | Journal of Medical Systems | Germany | ChatGPT-3.5, GPT-4 | Evaluation of ChatGPT and GPT-4 in pediatric emergencies: good diagnostic accuracy, limited emergency call advice, and first aid instruction, potential tools for lay responders and professionals. |
| Savage et al. [73] | 2023 | Original Research | JMIR Medical Informatics | USA | BioMed-RoBERTa | Development and validation of a language model-based screening tool for optimizing Best Practice Alerts in healthcare; use of AI to enhance EHR. |
| Goodman et al. [74] | 2023 | Original Investigation | JAMA Network Open | USA | ChatGPT-3.5, GPT-4 | Use of chatbots in medicine, specifically the accuracy and reliability of LLMs in responding to physician-generated medical questions. |
| Klang et al. [75] | 2023 | Original Research | BMC Medical Education | Israel | GPT-4 | Utilization of LLMs for creating medical examination questions; exploration of the benefits and limitations of LLMs in medical education. |
| Levin et al. [12] | 2024 | Original Research | International Journal of Nursing Studies | Israel | GPT-4, Claude-2 | Comparison of LLMs to neonatal nurses in clinical decision support for neonatal care. |
| Salvagno et al. [76] | 2023 | Perspective | Critical Care | Belgium | ChatGPT-3.5 | The use of LLMs in scientific writing; its potential applications, ethical considerations, and the need for proper regulations. |
| Pham et al. [77] | 2024 | Original Research | Journal of Medical Internet Research | USA | ChatGPT-3.5, GPT-4 | The performance of LLMs in simulations for cardiac arrest and bradycardia based on the American Heart Association's Advanced Cardiovascular Life Support guidelines. |
| Cabral et al. [78] | 2024 | Letter | JAMA Internal Medicine | USA | GPT-4 | Comparison of clinical reasoning capabilities between an LLM and human physicians (internal medicine residents and attending physicians). |
| Huespe et al. [79] | 2023 | Observational Study | Critical Care Explorations | Argentina | ChatGPT-3.5 | Evaluation of the capabilities of LLMs in generating the background sections of critical care clinical research questions compared to human researchers with different H-indices. |
| Si et al. [80] | 2019 | Original Research | Journal of the American Medical Informatics Association | USA | ELMo, BERT | Application of contextual embeddings (ELMo, BERT) to enhance clinical concept extraction from medical texts. |
| Almazyad et al. [81] | 2023 | Original Research | Cureus | Saudi Arabia | GPT-4 | The use of LLMs to enhance expert panel discussions at a medical conference, focusing on pediatric palliative care and ethical decision-making scenarios. |
| Guillen-Grima et al. [82] | 2023 | Original Research | Clinics and Practice | Spain | ChatGPT-3.5, GPT-4 | Use of LLMs to answer the Spanish Medical Residency Entrance Examination; evaluation of AI's ability to navigate complex medical exams. |
| Chung et al. [83] | 2024 | Original Research | JAMA surgery | USA | GPT-4 Turbo | Evaluating the performance of LLMs in perioperative risk stratification and prognostication across various tasks using EHR data. |
| Abdullahi et al. [84] | 2024 | Original Research | JMIR Medical Education | USA | Bard, ChatGPT-3.5, GPT-4 | The effectiveness of LLMs in diagnosing rare and complex medical conditions, focusing on improving medical education and diagnostic accuracy. |
| Benboujja et al. [85] | 2024 | Original Research | Frontiers in Public Health | USA | GPT-4 | Development and application of a multilingual, AI-driven educational curriculum in pediatric care to overcome language barriers in global healthcare education. |
| Igarashi et al. [86] | 2024 | Original Research | Journal of Nippon Medical School | Japan | GPT-4 | Evaluation of an LLM on emergency medicine board certification examinations, focusing on the model's performance without the use of visual aids. |
| Tran et al. [87] | 2024 | Original Research | Journal of Surgical Research | USA | ChatGPT-3.5 | Evaluation of an LLM's performance on general surgery in-training examination questions, exploring its capabilities in surgical education. |
| Amacher et al. [88] | 2024 | Original Research | Resuscitation Plus | Switzerland | GPT-4 | Evaluating the prognostic accuracy of LLMs in predicting death and poor neurological outcomes at hospital discharge among cardiac arrest patients. |

| Doshi et al. [89] | 2024 | Original Research | Radiology | USA | ChatGPT-3.5, GPT-4, Bard (Gemini), Bing | Evaluation of LLMs in simplifying radiology report impressions to improve readability and patient understanding. |
| Yoon et al. [90] | 2024 | Original Research | Journal of Medical Internet Research | Korea | ChatGPT-3.5 | Exploring the role of LLMs in enhancing health care data interoperability by facilitating data transformation and exchange. |
| Frosolini et al. [91] | 2024 | Original Research | Diagnostics | Italy | GPT-4, Gemini | Evaluation of LLMs for triaging maxillofacial trauma cases, comparing the performance against traditional methods. |
| Islam et al. [92] | 2023 | Original Research | IEEE International Conference on Big Data | USA | GPT-4, BioGPT | Development of an autocompletion tool for Chief Complaints in EHR using LLMs to enhance documentation efficiency in Emergency Departments. |
| Lu et al. [1] | 2023 | Letter | Annals of Biomedical Engineering | China | ChatGPT-3.5, GPT-4 | Exploration of potential uses of LLMs in Intensive Care Medicine, focusing on knowledge augmentation, device management, clinical decision support, early warning systems, and ICU database establishment. |
| Madden et al. [93] | 2023 | Letter | Intensive Care Medicine | Ireland | GPT-4 | Evaluation of the effectiveness of LLMs in querying and summarizing unstructured medical notes in the ICU setting. |

## 4.2 Bibliometric Analysis

### 4.2.1 Key Characteristics of Literature

This scoping review included a focused selection of 24 articles, providing a global perspective on applying LLMs in CCM. This diverse corpus spans several countries, demonstrating widespread research interest in applying LLMs in CCM.

**Country** The distribution of the selected publications indicates substantial international collaboration and research efforts. We analyzed the countries where each selected article's first and corresponding authors were based. The authors from the United States took the lead in most studies (n=11, 45.8%), followed by authors from Israel (n=2, 8.33%), and individuals from Argentina, Belgium, China, Germany, Ireland, Italy, Japan, Korea, Saudi Arabia, Spain, and Switzerland, each contributing one article. It revealed that nearly half of the studies were conducted in the United States, with much fewer contributions from other countries and regions. This indicates a concentration of research activities in applying LLMs in CCM within the United States, potentially reflecting the advanced development and adoption of AI technologies in American critical care settings.

**Article Type** Regarding the type of publications, most were original research articles (n=20, 83.3%), reflecting strong empirical research interest in applying LLMs in CCM. Three letters (12.5%) and one perspective article (4.2%) supplemented this.

**LLMs in Utilization** GPT-4 appears most frequently, cited in 18 articles (75%), demonstrating its relevance and recent prominence in CCM applications. Other models include ChatGPT-3.5 (n=11, 45.8%) and models like Bard, Gemini, and Claude-2, highlighting the breadth of generative models explored in the included studies.

### 4.2.2 Keyword Co-occurrence Network

As shown in Figure 3, 131 unique keywords were extracted from the literature using the VOSviewer software, among which "artificial intelligence" emerged as the most frequently occurring term, appearing in 9 articles with a total link strength of 98. This highlights the central role of AI in the reviewed studies. The keyword "ChatGPT" also featured prominently, with six occurrences and a total link strength of 61, indicating a strong research interest in the capabilities or potentials of this specific model. Other notable keywords include "GPT-4" and "clinical decision support", each with unique roles in the evolving landscape of LLMs in CCM. This underlines a significant gap in research that specifically targets the integration of LLMs within critical care settings.

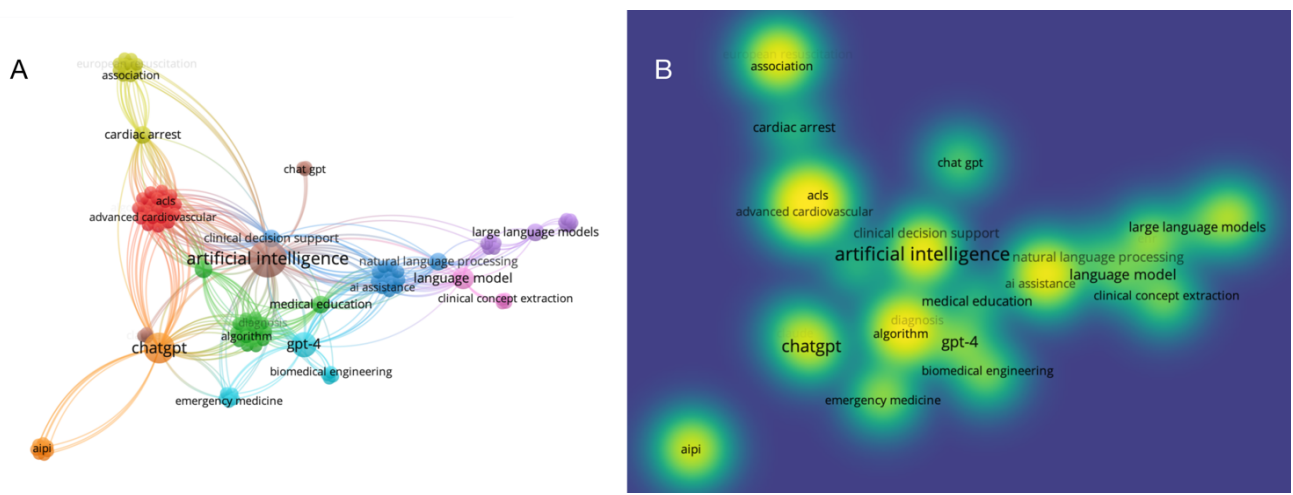

**Figure 3. Keyword co-occurrence network analysis using the VOSviewer software. (A) Network Visualization, (B) Density Visualization.**

## 4.3 Applications of LLMs in CCM

### 4.3.1 Clinical Decision Support

This section discusses the focus of LLMs in CCM for clinical decision support, focusing on their involvement in prehospital and in-hospital critical care, as illustrated in Figure 4.

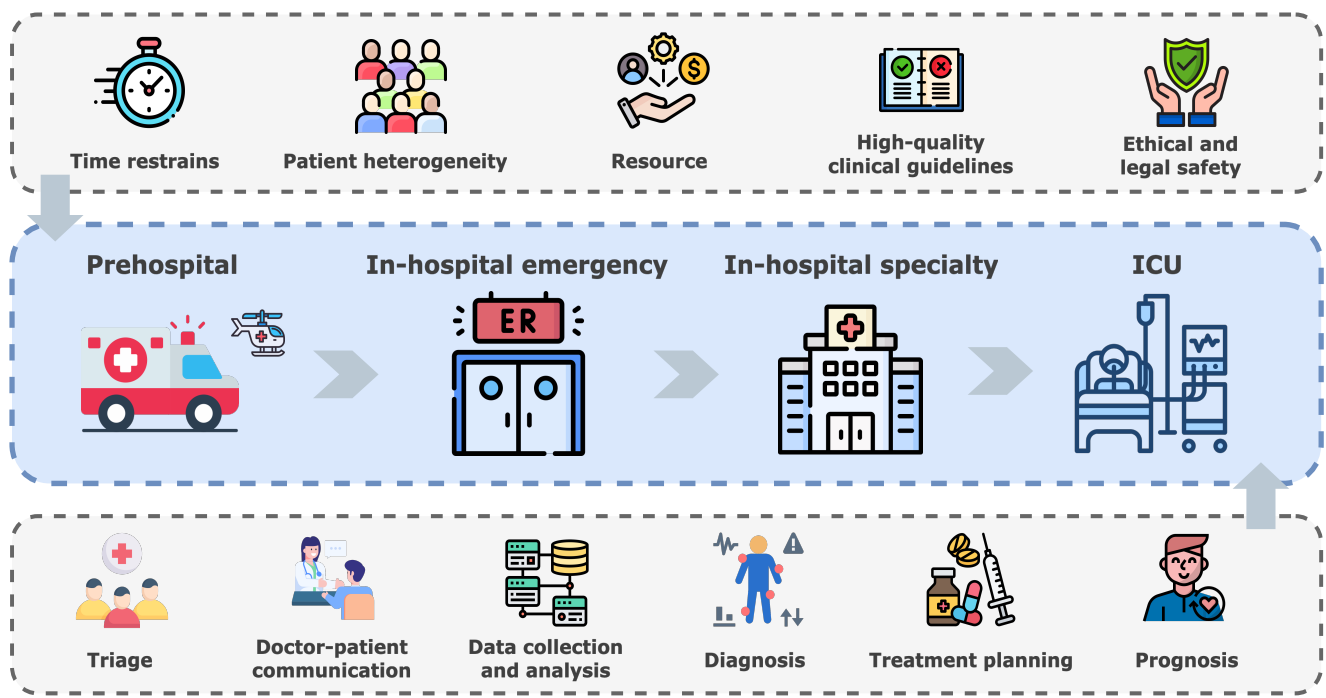

**Figure 4. The process and characteristics of clinical decision support in prehospital and in-hospital critical care.**

**Prehospital** During the prehospital phase, LLMs can assist pre-hospital rescuers in providing patient education and clinical risk stratification. Bushuven et al. [72] employed ChatGPT-3.5 and GPT-4 to support clinical decision-making in prehospital basic life support and pediatric advanced life support scenarios using 22 case vignettes that were developed and validated by five emergency physicians. The primary purpose of LLMs in their study was to recognize emergencies and offer valid advice to lay rescuers. It was found that although both models correctly identified the emergency in most cases (94%), they only advised calling emergency services in 54% of the scenarios. Additionally, LLM-provided first aid instructions were correct in 45% of the cases, while 3 of 22 cases (13.6%) LLM-generated advanced life support techniques were incorrect. These findings indicated that the validity, reliability, and safety of ChatGPT/GPT-4 as an emergency support tool were questionable. However, whether humans would perform better in the same situation was uncertain, and further evaluation of LLMs' context sensitivity and adaptability is needed.

LLMs can be applied in triage, diagnosis, treatment planning, and prognosis prediction in in-hospital critical care settings.

**Triage** Frosolini et al. [91] assessed the feasibility of LLMs in triaging complex maxillofacial trauma cases. In their study, patient records from a tertiary referral center were used, and the triage suggestions provided by ChatGPT-4 and Gemini were compared against the real-world recommendations at the study center. They found that the recommendations given by the LLMs were generally consistent with those given by the referral center, with ChatGPT achieving higher consistency in treatment recommendations and specialist consultations compared to Gemini.

**Diagnosis** LLMs demonstrate the potential to aid physicians in making diagnostic decisions. Abdullahi et al. [84] used three popular LLMs, Bard, ChatGPT-3.5, and GPT-4, to explore their potential to aid the diagnosis of rare and complex diseases. The study utilized prompt engineering techniques on publicly available diagnostic cases, comparing the three LLMs' performance against human respondents and a medical-specific LLM, MedAlpaca. The findings revealed that the three studied LLMs outperformed the average human consensus and MedAlpaca in diagnostic accuracy through prompt engineering.

**Treatment Planning** In treatment planning, LLMs show considerable promise in providing personalized treatment recommendations and optimizing clinical pathways. Savage et al. [73] developed and validated the LLM screening tool to selectively identify patients appropriate for deep vein thrombosis anticoagulation prophylaxis best practice alerts (BPAs) in EHRs. They found that the LLM screening tool improved the precision of BPAs, reducing the number of unnecessary alerts by 20% and increasing the applicability of alerts by 14.8%. Pham et al. [77] evaluated ChatGPT's performance in treating cardiac arrest and bradycardia simulations following the American Heart Association's Advanced Cardiovascular Life Support (ACLS) guidelines. Using the 2020 ACLS guidelines, ChatGPT's responses to two simulation scenarios were assessed for accuracy. The study found that ChatGPT had a median accuracy of 69% for cardiac arrest and 42% for bradycardia, with significant variability in its outputs, often missing critical actions, and having incorrect medication information. This study highlighted the need for consistent and reliable guidance to prevent potential medical errors and optimize the application of LLM to improve its reliability and effectiveness in clinical practice.

**Prognosis Prediction** Amacher et al. [88] used GPT-4 to predict mortality and poor neurological outcomes at hospital discharge for adult cardiac arrest patients. The study involved prompting GPT-4 with sixteen prognostic parameters from established post-cardiac arrest scores. The findings showed that GPT-4 achieved an AUC of 0.85 for in-hospital mortality and 0.84 for poor neurological outcomes, comparable to traditional scoring systems. Despite these promising results, the study highlighted the need for human oversight due to instances of illogical answers provided by the model. It indicated the model's potential to provide valuable prognostic insights. Chung et al. [83] used GPT-4 to perform risk stratification and predict postoperative outcomes based on procedure descriptions and preoperative clinical notes from EHRs. They found that GPT-4 achieved F1 scores of 0.64 for hospital admission, 0.81 for ICU admission, 0.61 for unplanned admission, and 0.86 for hospital mortality prediction. However, the model performed poorly in predicting numerical outcomes, indicating that while LLMs may aid in perioperative risk stratification, they are inadequate for numerical predictions.

### 4.3.2 Medical Documentation and Reporting

LLMs are making strides in medical documentation and reporting by automatizing and streamlining these processes. Doshi et al. [89] used four LLMs (ChatGPT-3.5, GPT-4, Bard, and Bing) to simplify the process of producing radiology report impressions (RRI). Utilizing 750 anonymized RRI, the study employed three distinct prompts to evaluate the improvement in readability. The results showed that all four LLMs simplified the RRI generation process across various imaging modalities and prompts. Although all models improved readability, their effectiveness varied based on prompt wording. Islam et al. [92] employed LLMs to create an autocompletion tool for chief complaints in emergency department settings. This tool aims to expedite the documentation process, enabling patient complaints to be recorded accurately and timely. Salvagno et al. [76] assessed the utility of ChatGPT in organizing material, generating drafts, and proofreading scientific texts. They found that while ChatGPT could efficiently produce initial drafts and provide language edits, the output often required substantial revision by human experts to ensure accuracy and relevancy.

Yoon et al. [90] explored the role of LLMs in enhancing healthcare data interoperability by using ChatGPT-3.5 for information exchange tasks. The results showed that LLMs can significantly improve the interoperability of data without the need for complex standardization processes. Si et al. [80] explored the impact of ELMo and BERT on clinical concept extraction tasks using data from the Medical Information Mart for Intensive Care (MIMIC) III and other clinical corpora. They found that contextual embeddings pre-trained on a large clinical corpus significantly outperformed traditional methods. This demonstrates the potential of contextual embeddings in encoding valuable semantic information for clinical NLP tasks. Madden et al. [93] used ChatGPT-4 to query and summarize unstructured medical notes in the ICU. They found that while the model could produce concise and useful summaries, it also had significant risks of generating hallucinations. Additionally, Okada et al. [94] explored the potential of LLMs in assisting regulatory reporting by summarizing critical information and providing clear and actionable insights. They found that LLMs could significantly improve the accuracy and efficiency of regulatory documentation in emergency scenarios.

### 4.3.3 Medical Education and Doctor-Patient Communication

LLMs are utilized more and more frequently in the medical education field now. One important area closely connected to LLMs is to generate or answer questions in the medical exam. Guillen-Grima et al. [82] evaluated the performance of GPT-3.5 and GPT-4 in navigating the Spanish Medical Residency Entrance Examination. The study gauged overall performance, analyzed discrepancies across medical specialties, and distinguished between theoretical and practical questions. The results showed that GPT-4 significantly outperformed GPT-3.5, achieving an 86.81% correct response rate in Spanish and performing slightly better with English translations. Igarashi et al. [86] assessed the performance of GPT-4 on the Japanese Emergency Medicine Board Certification Examinations over five years. The LLM was tasked with answering text-based and scenario-based questions, achieving an overall correct response rate of 62.3% and a substantial agreement rate (kappa = 0.70) between repeated responses. Tran et al. [87] tested the performance of ChatGPT-3.5 on the American Board of Surgery In-Training Examination preparation questions. Using 200 randomly selected general surgery multiple-choice questions, ChatGPT achieved a correct response rate of 62%. Klang et al. [75] utilized GPT-4 to generate multiple-choice questions for medical examinations 2023. The AI tool was tasked with creating a 210-question exam based on an existing template, and specialist physicians evaluated the generated questions. The study found that GPT-4 was efficient, with only one question (0.5%) being completely incorrect and 15% requiring revisions.

Meanwhile, LLMs use information such as clinical guidelines to answer questions from real-world medical scenarios. Cabral et al. [78] compared LLMs to physicians in processing medical data and clinical reasoning, and they found that LLMs outperformed doctors in data processing and reasoning. Goodman et al. [74] assessed the accuracy and reliability of ChatGPT-generated responses to 284 physician-generated medical questions. They found that ChatGPT provided predominantly accurate information, with a median accuracy score of 5.5 (between almost completely and completely correct) and a median completeness score of 3.0 (complete and comprehensive). Levin et al. [12] employed two LLMs, GPT-4 and Claude-2.0, to provide initial assessment and treatment recommendations for patients in neonatal intensive care settings. The results indicated that both models demonstrated clinical reasoning abilities, with Claude-2.0 outperforming GPT-4 in clinical accuracy and response speed.

LLMs also can be used to overcome language barriers and enhance communication. Benboujja et al. [85] developed and evaluated a multilingual, AI-driven curriculum to overcome language barriers in pediatric care. Using GPT-4 for translation, the study created 45 educational video modules in English and Spanish, covering surgical procedures, perioperative care, and patient journeys. Almazyad et al. [81] employed GPT-4 to enhance expert panel discussions in pediatric palliative care. They found that GPT-4 effectively facilitated discussions on do-not-resuscitate conflicts by summarizing key themes such as communication, collaboration, patient and family-centered care, trust, and ethical considerations.

## 5. Discussion

With the recent advent of LLMs, the field of CCM has witnessed groundbreaking developments and advancements [95]. At present, there are many review articles focusing on the application of LLMs in health and medicine [16-18], and there are also many review articles on AI in CCM [65-67], but few review articles focusing on the application of LLMs in CCM, except for some editorials and letters [1, 96]. The primary objective of this scoping review is to provide a comprehensive portrait of the applications of LLMs in CCM, identifying the advantages, challenges, and future research directions of this area. An extensive examination of the selected literature revealed that LLMs hold substantial promise for transforming clinical practice and improving patient outcomes in CCM. The current applications, together with the challenges and future directions of LLMs in CCM identified by this review, are shown in Figure 5.

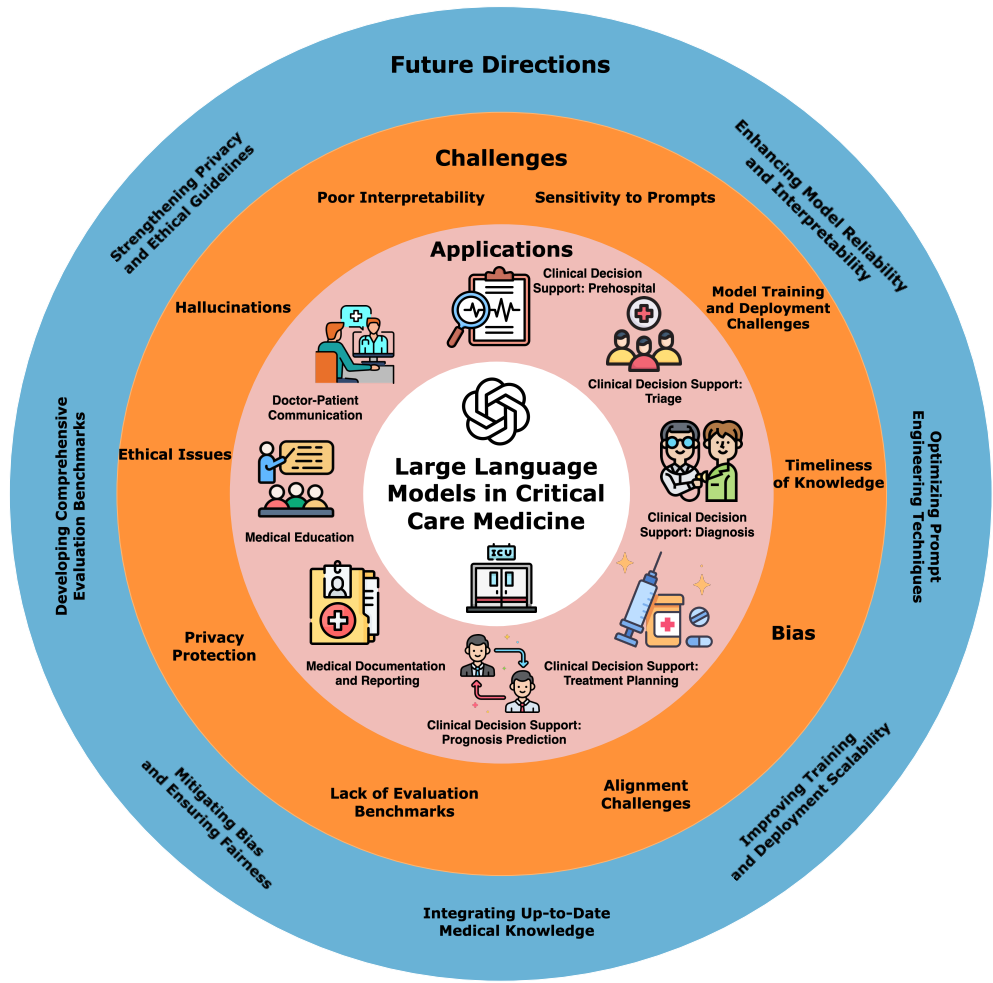

**Figure 5. The applications, challenges, and future directions of LLMs in CCM.**

## 5.1 Advantages

The application of LLMs in CCM has demonstrated numerous advantages. Compared to traditional machine learning techniques, using LLM technology in CCM reveals that LLMs can effectively understand and generate natural language, which can aid clinicians in writing patient medical records and diagnostic notes [76, 89, 92, 97]. The capabilities of LLMs extend beyond text interpretation and generation. They surpass traditional machine learning methods in unstructured data handling. LLMs can learn directly from extensive patient data without manual feature engineering. Moreover, multimodal LLMs can learn and understand medical images, such as X-rays and CT scans [18].

In clinical practice, LLMs can extract critical information from a patient's historical medical records and combine it with the latest medical research, aiding physicians in identifying rare diseases or those with early symptoms that are not clearly defined [98]. For medical research, LLMs can assist researchers in summarizing data and information in literature research and providing suggestions for manuscript structure, references, and titles, enhancing the readability and completeness of texts [76]. LLMs have a wide range of knowledge so that they can provide physicians with a comprehensive analysis for decision-making across different specialties [97]. Integrating LLMs into CCM represents a paradigm shift with the potential to reshape critical care profoundly [82].

## 5.3 Challenges

Currently, LLMs face some challenges in their applications in the field of CCM.

**Hallucinations and Poor Interpretability** LLMs may produce hallucinations or generate information that does not correspond with facts [99]. In CCM, this may pose a high risk to patient outcomes as incorrect recommendations may lead to incorrect diagnosis or treatment plans, potentially endangering patient lives. Also, the decision-making processes of LLMs are typically opaque [17], making it difficult for users to track and understand how the models extract features from input data and generate recommendations [100]. The lack of transparency about the sources of factual information further complicates the application of LLMs in clinical decision-making [74].

**Sensitivity to Prompts** LLM-generated outputs are highly sensitive to input prompts, and different prompt strategies may affect the model's capabilities and performance [72]. The variability in prompts may lead to inconsistent results, necessitating the analysis of multiple iterations and varied prompts to ensure the accuracy and reliability of LLM applications in CCM. There is no one-size-fits-all prompting approach that can improve the performance of LLMs, and a single strategy is not universally applicable to all LLMs [85].

**Model Training and Deployment Challenges** Training and deploying LLMs require substantial computational resources and time. High precision and reliability are necessary in medical applications, requiring models to be trained on large, diverse datasets. However, the availability of large-volume data in CCM is limited, with public databases like MIMIC [101, 102] and eICU Collaborative Research Database [103] commonly used for pre-training large models. Strict hospital regulations often restrict data sharing, complicating multicenter studies [104, 105]. Additionally, LLMs may not comply with data privacy standards required in local computing environments, which restricts their training and deployment in critical care settings [73].

**Timeliness of Knowledge** CCM is a rapidly progressing field with the continuous emergence of new treatments and medical discoveries. LLMs may lack the capability to update synchronously with the latest critical care research, impacting their performance in clinical practice [13].

**Bias and Alignment Challenges** LLMs may unintentionally learn biases from the training data and reproduce them in their outputs. This could lead to biased treatment recommendations for certain patient groups, affecting the quality of medical care [106]. Also, the inherent biases in LLM selection remain a concern [72]. It may lead to disseminating inaccurate model recommendations, potentially jeopardizing patient safety [74]. The models' decision-making processes and outputs must align with clinical guidelines and medical ethics [107]. Moreover, as clinical practice involves complex human behaviors and non-standardized decision-making processes, aligning the behavior of LLMs is crucial for the models to interpret clinical data correctly and respond appropriately [108].

**Lack of Evaluation Benchmarks** In medicine, there is a lack of unified and widely accepted standards for evaluating the performance of LLMs. This makes it difficult to assess and compare the effectiveness of different models and accurately measure their specific impact on clinical practice [109]. Traditional model evaluation mainly focuses on the accuracy of medical question answering, which cannot fully reflect the capabilities of LLMs in clinical practice [110]. It is important to establish specific evaluation frameworks and performance metrics [108].

**Privacy Protection and Ethical Issues** In CCM, LLMs must process vast amounts of sensitive data with patient privacy. The lack of effective supervision mechanisms may cause models to deviate from established treatment directions, resulting in unpredictable behaviors and increasing the risk of medical errors and violation of medical ethics. Compliance with contemporary privacy laws when handling patient information is imperative to avoid legal pitfalls [111]. Furthermore, ethical issues must be addressed to ensure responsible LLM use in critical care environments [1, 81].

## 5.4 Future Directions

The future development of LLMs in CCM should focus on the following key areas.

**Enhancing Model Reliability and Interpretability** The accuracy and reliability of LLMs can be enhanced by improving training data quality, employing ensemble learning, and implementing adversarial training [112-114]. Developing and fine-tuning LLMs specifically for CCM, incorporating domain-specific knowledge, are also essential measures to improve model accuracy and reliability. Future research should also focus on increasing the use of plugins to evaluate results, which can increase the reliability of models by reducing the generation of erroneous results and references by checking results with external knowledge from databases such as PubMed [88]. Using attribution methods as post-hoc explanations for LLMs, researchers can have a deeper understanding of the operating mechanism of LLMs [115]. In addition, the interpretability of LLM can also be improved through CoT, tree-of-thoughts, graph-of-thoughts, and retrieval augmented generation (RAG) technologies [115, 116].

**Optimizing Prompt Engineering Techniques** LLMs' sensitivity to prompts necessitates continuous model training and updates to ensure accuracy. Recently, a prompt engineering framework called Medprompt developed by Microsoft uses a combination of three main components (dynamic few-shot, self-generated CoT, and choice shuffle ensemble) to combine powerful general LLMs (such as GPT-4) with effective prompt engineering, outperforming LLMs that fine-tuned on domain-specific data [117]. Future research needs to investigate more robust prompt engineering techniques, including CoT, few shots, and RAG, to enhance the consistency and reliability of LLM outputs in different clinical scenarios.

**Improving Training and Deployment Scalability** To address LLM training and deployment challenges, scalable model architectures, transfer learning, model pruning, and federated learning approaches can be explored to reduce computational demands and facilitate practical deployment [118]. The emergence of low-powered open-source LLMs running locally could circumvent issues related to data privacy and computational resource constraints [93]. It is crucial to convert medical datasets into easily accessible structured databases and train healthcare professionals to utilize LLMs in clinical practice to aid decision-making [88].

**Integrating Up-to-Date Medical Knowledge** Employing online learning systems allows models to update and assimilate the latest medical research and changes in clinical practices in a timely manner. Additionally, modular update systems can swiftly integrate new medical discoveries, while expert collaboration ensures the scientific validity and timeliness of model outputs. Moreover, using RAG techniques to connect LLMs with databases in CCM can also address the knowledge timeliness issue to some extent [116].

**Mitigating Bias and Ensuring Fairness** Bias mitigation should be approached through pre-processing, in-training, intra-processing, and post-processing stages [106]. Pre-processing techniques involve modifying model inputs to ensure balanced representations. In-training methods focus on adjusting model parameters to mitigate biases through gradient-based updates. Intra-processing methods modify inference behavior without further training, while post-processing techniques correct model outputs to ensure fair treatment across demographic groups. Developing bias detection and dataset augmentation algorithms to review and adjust model outputs regularly can help reduce model bias and ensure fairness [119].

**Developing Comprehensive Evaluation Benchmarks** Recent studies demonstrated that performance varies significantly across different medical tasks, highlighting the need for task-specific evaluations [110, 120]. Future directions should focus on developing multi-dimensional evaluation benchmarks that go beyond accuracy to include factors such as clinical relevance, interpretability, and robustness under real-world conditions and task-specific benchmarks that focus on the specific area of CCM. Collaboration between medical professionals and AI researchers is crucial to designing these benchmarks, ensuring they are clinically meaningful and practically applicable [110]. Establishing such standards will facilitate more reliable comparisons between models, guiding improvements and fostering trust in their deployment in CCM [110].

**Strengthening Privacy and Ethical Guidelines** Efforts need to be made to establish flexible and robust standards and regulations to safely and effectively implement LLMs in CCM [74]. Employing synthetic data generation techniques to expand training datasets can support extensive and effective model training while protecting patient privacy [121]. Strengthening collaboration with policymakers, ethicists, and legal experts is necessary to ensure LLM applications comply with ethical and legal requirements, thus protecting patient privacy and data security [122].

# 6. Conclusion

In conclusion, although LLMs in CCM are not yet ICU experts, they act more than stochastic parrots. Applying LLMs in CCM presents a transformative potential for enhancing critical care. Through techniques such as RAG and working as agents to integrate with external knowledge bases, LLMs can learn and acquire professional and up-to-date knowledge relevant to CCM. LLMs are capable of reasoning beyond random generation, and they have demonstrated capabilities to improve diagnostic accuracy, plan treatments, and provide valuable support in prognosis prediction. However, applying LLMs in CCM is still in its early stages, with few large models specifically designed and fine-tuned for this domain. Future research should focus on enhancing model reliability and interpretability, optimizing prompt engineering techniques, improving training and deployment scalability, integrating up-to-date medical knowledge, mitigating bias and ensuring fairness, developing comprehensive evaluation benchmarks, and strengthening privacy and ethical guidelines. Close collaboration across multiple disciplines, such as medicine, computer science, and data science, can catalyze applying LLMs in CCM. There is some way to go before making LLMs become true ICU experts. Nevertheless, we are optimistic that LLMs in CCM will become experts in the near future, helping to improve the quality of critical care and the outcomes of critically ill patients.

## Acknowledgments

This work was supported by Grants from the Zhejiang Provincial Natural Science Foundation of China (No. LZ22F020014), the National Natural Science Foundation of China (No. 82372095), the Beijing Natural Science Foundation of China (No. 7212201, QY23066), the Humanities and Social Science Project of Chinese Ministry of Education (No. 22YJA630036).

## Author Contributions

T.S. and G.K. conceived and designed the study. T.S., Z.Y., and G.K. conducted the literature review and analyzed data. T.S., J.M., Z.Y., H.X., MQ.X., MR.X., Y.L., and G.K. drafted the manuscript. J.M. and H.Z. provided clinical support and advice. T.S. and G.K. revised the manuscript. All authors contributed to the interpretation and preparation of the final manuscript. All authors read and approved the final manuscript.

## Competing Interests

The authors declare no competing interests.

Shi et al. 2024